\def\year{2022}\relax
\documentclass[letterpaper]{article} 
\usepackage{aaai22}  
\usepackage{times}  
\usepackage{helvet}  
\usepackage{courier}  
\usepackage[hyphens]{url}  
\usepackage{graphicx} 
\usepackage{natbib}  
\usepackage{caption} 
\DeclareCaptionStyle{ruled}{labelfont=normalfont,labelsep=colon,strut=off} 
\usepackage{algorithm}
\usepackage{algorithmic}
\usepackage[switch]{lineno}
\usepackage{booktabs}
\usepackage{threeparttable}
\usepackage{multirow}
\usepackage{amssymb}
\usepackage{subfigure}
\usepackage{newfloat}
\usepackage{listings}
\usepackage{marvosym}
\floatstyle{ruled}
\newfloat{listing}{tb}{lst}{}
\floatname{listing}{Listing}
\title{Towards Ultra-Resolution Neural Style Transfer via\\ Thumbnail Instance Normalization}
\author{
    Zhe Chen\textsuperscript{\rm 1},
    Wenhai Wang\textsuperscript{\rm 2}\textsuperscript{\Letter},
    Enze Xie\textsuperscript{\rm 3},
	Tong Lu\textsuperscript{\rm 1}\textsuperscript{\Letter},
	Ping Luo\textsuperscript{\rm 3}\\
}
\affiliations{
	$^1$State Key Lab for Novel Software Technology, Nanjing University~~~\\
	$^2$Shanghai Artificial Intelligence Laboratory~~~
	$^3$The University of Hong Kong~~~\\
	chenzhe98@smail.nju.edu.cn,
	wangwenhai@pjlab.org.cn,
	xieenze@hku.hk, 
	lutong@nju.edu.cn, 
	pluo@cs.hku.hk
}

\def\ie{\emph{i.e.}}
\def\eg{\emph{e.g.}}

\def\etal{{\em et al.}}

\usepackage{bibentry}

\newcommand\blfootnote[1]{%
\begingroup
\renewcommand\thefootnote{}\footnote{#1}%
\addtocounter{footnote}{-1}%
\endgroup
}

\begin{document}
\maketitle

\begin{abstract}
We present an extremely simple \textbf{U}ltra-\textbf{R}esolution \textbf{S}tyle \textbf{T}ransfer framework, termed URST, to flexibly process arbitrary high-resolution images~(\eg, 10000$\times$10000 pixels) style transfer for the first time.
Most of the existing state-of-the-art methods would fall short due to massive memory cost and small stroke size when processing ultra-high resolution images.
URST completely avoids the memory problem caused by ultra-high resolution images by (1) dividing the image into small patches and (2) performing patch-wise style transfer with a novel Thumbnail Instance Normalization (TIN).
Specifically, TIN can extract thumbnail features' normalization statistics and apply them to small patches, ensuring the style consistency among different patches.

Overall, the URST framework has three merits compared to prior arts. 
(1) We divide input image into small patches and adopt TIN, successfully transferring image style with arbitrary high-resolution.
(2) Experiments show that our URST surpasses existing SOTA methods on ultra-high resolution images benefiting from the effectiveness of the proposed stroke perceptual loss in enlarging the stroke size.
(3) Our URST can be easily plugged into most existing style transfer methods and directly improve their performance even without training.
Code is available at \url{https://git.io/URST}.
\blfootnote{\Letter~Corresponding authors.}
\end{abstract}

\section{Introduction}
\label{sec:introduction}

With the development of deep learning, neural style transfer has achieved remarkable success~\cite{johnson2016perceptual,lu2017decoder,shen2018neural,sanakoyeu2018style,li2019learning}, but ultra-high resolution style transfer is rarely explored in these works.
In natural scenes, ultra-high resolution images are often seen in large posters, photography works, and ultra-high definition (\eg, 8K) videos. 
There are two main challenges when stylizing ultra-high resolution images: 
(1) The massive memory cost of ultra-high resolution images may exceed the GPU memory capacity.
(2) Small stroke size may cause unpleasant dense textures in ultra-high resolution results.

\begin{figure}[t]
	\centering
	\includegraphics[width=0.47\textwidth]{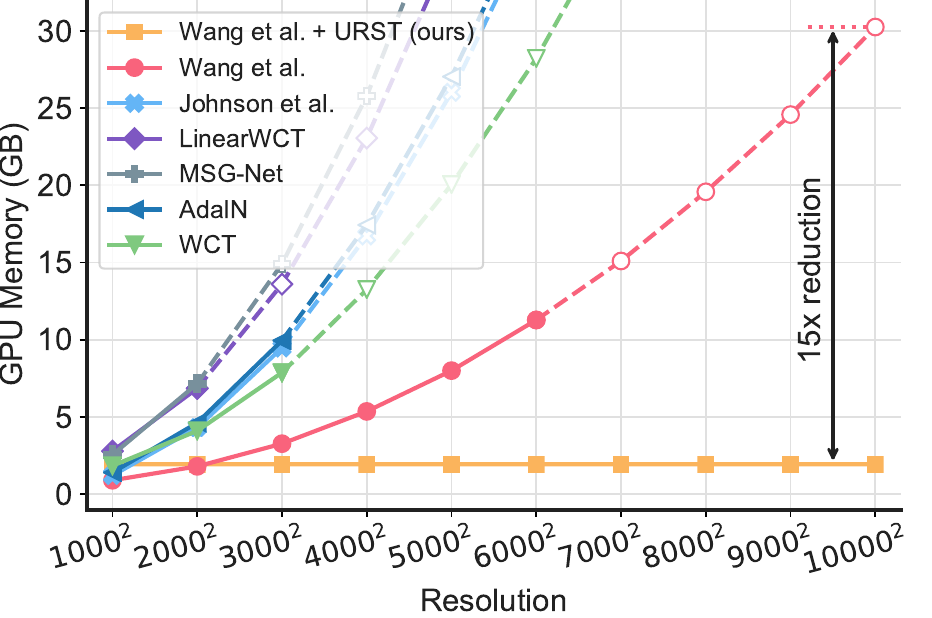}
	\caption{\textbf{GPU memory comparison of different style transfer methods.}
	The hollow markers indicate that images of these resolutions cannot be rendered on a 12GB GPU (Titan XP) due to memory limitation.
	By contrast, our URST not only can process images of unconstrained resolution, but also achieves a 15$\times$ reduction in memory when stylizing an ultra-high resolution image of 10000$\times$10000 pixels by using the model of \cite{wang2020collaborative}. 
	}
	\label{fig:comparison_memory} 
\end{figure}

First, for the memory limitation, the existing methods mainly use lightweight network architecture~\cite{jing2020dynamic}, model pruning~\cite{an2020real}, and knowledge distillation~\cite{wang2020collaborative} to reduce memory cost. However, most of these methods are palliatives.
As shown in Figure~\ref{fig:comparison_memory}, with the growth of the input resolution, the memory cost of the distillation-based method \cite{wang2020collaborative} increases sharply and finally runs out of the GPU memory (12GB in Titan XP).
This phenomenon motivates us to design a more effective strategy for stylizing ultra-high resolution images.

\begin{figure*}[ht]
	\centering
	\includegraphics[width=0.99\textwidth]{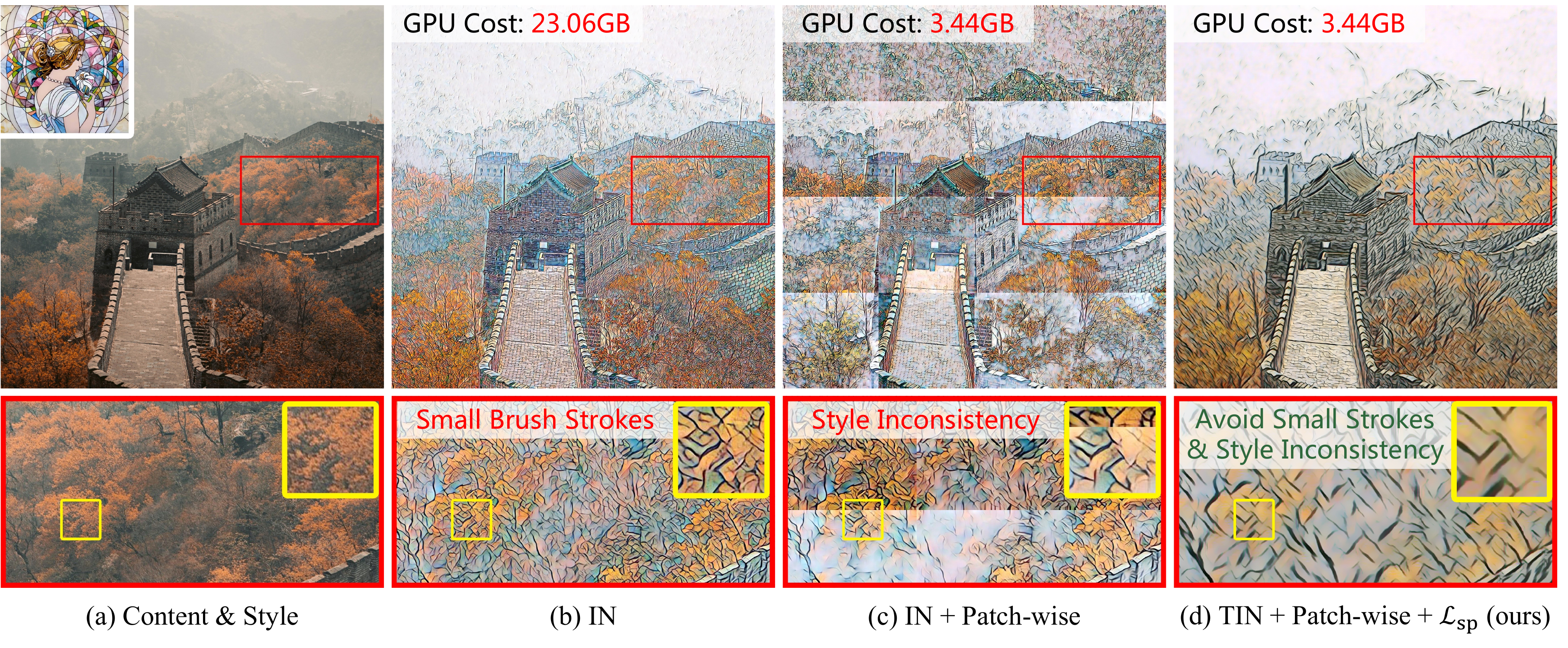}
	\caption{\textbf{Ultra-high resolution stylized results (4000$\times$4000 pixels) generated by different methods.} 
	(a) Content image and style image. 
	(b) Result produced by the original instance normalization (IN), which costs 23.06GB GPU memory. Compared with our result (d), its brush strokes are relatively small. 
	(c) Result produced by patch-wise style transfer with IN, which only costs 3.44GB, but its stylized patches are inconsistent in style. 
	(d) Result produced by patch-wise style transfer with the proposed thumbnail instance normalization (TIN) and stroke perceptual loss $\mathcal{L}_{\rm sp}$, which also costs 3.44GB. 
	Benefiting from these two designs, we can obtain ultra-high resolution stylized results with large brush strokes under limited memory resources.
	}
	\label{fig:comparison_linearwct} 
\end{figure*}

The second problem is that the brush strokes in ultra-high resolution stylized results are relatively small. 
As shown in Figure~\ref{fig:comparison_linearwct}(b), when given a high-resolution input, the model with small brush strokes would produce unpleasant dense textures. 
Enlarging the stroke size is a widely-used approach to address this problem. At present, the existing methods can be roughly divided into two categories. 
One is to train or inference with large style images~\cite{jing2019neural, li2017universal, zhang2018multi}. Another solution is to enlarge the receptive field of the style transfer network~\cite{jing2018stroke, wang2017multimodal}.
However, most of these methods tend to take extra inference time and memory, are not suitable for ultra-high resolution style transfer.

To compensate the above limitations, this work proposes an \textbf{U}ltra-\textbf{R}esolution \textbf{S}tyle \textbf{T}ransfer framework, termed URST.
Different from previous methods~\cite{jing2018stroke,an2020real, wang2020collaborative, wang2017multimodal}, our method 
(1) takes small patches instead of a full image as input, which makes it possible to process arbitrary high-resolution images under limited memory resources. 
(2) We replace the original instance normalization (IN)~\cite{ulyanov2016instance} by the proposed thumbnail instance normalization (TIN), to ensure the style consistency among different patches. As shown in Figure~\ref{fig:comparison_linearwct}(c), if we perform patch-wise style transfer with IN directly, style inconsistency among different patches would make them cannot be assembled into a pleasing image.
(3) We propose a stroke perceptual loss as an auxiliary loss for neural style transfer, motivating style transfer networks to keep large brush strokes.

Overall, the proposed URST has three advantages:

(1) Our framework can process arbitrary high-resolution images with limited memory. As shown in Figure~\ref{fig:comparison_memory}, when stylizing an ultra-high resolution image of 10000$\times$10000 pixels based on \cite{wang2020collaborative}, our framework only requires 1.94GB memory, while the original method needs 30.25GB, 15 times larger. To our knowledge, it is the first unconstrained resolution style transfer method.

(2) Our framework achieves high-quality style transfer of ultra-high resolution images. As shown in Figure~\ref{fig:comparison_linearwct}(d), our method uses larger brush strokes to depict the scene, which is much better than the effects presented in Figure~\ref{fig:comparison_linearwct}(b).

(3) Our framework can be easily plugged into most existing style transfer methods. Even without training, our framework can also obtain high-resolution results.

\begin{figure*}[ht]
	\centering
	\includegraphics[width=0.99\textwidth]{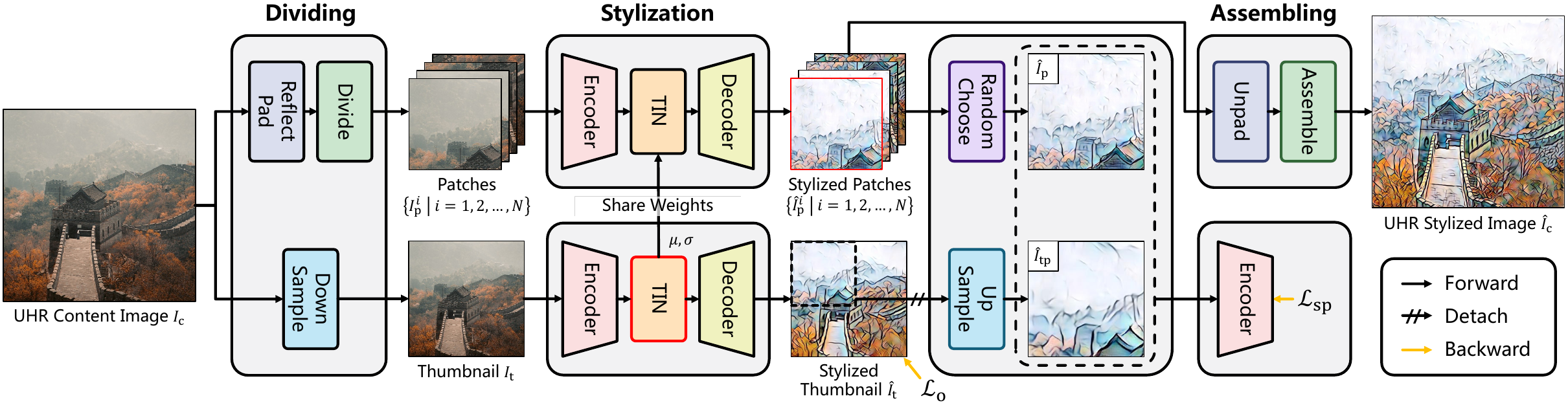}
	\caption{\textbf{Overall architecture of URST.}
	Its pipeline is divided into three stages: dividing, stylization, and assembling. 
	The core idea of URST is to divide the ultra-high resolution (UHR) content image into small patches and perform patch-wise style transfer with the proposed TIN. 
	The style transfer network in our framework can be different methods.
	In addition to the original loss $\mathcal{L}_{\rm o}$ of the selected method, our URST includes an auxiliary loss termed stroke perceptual loss $\mathcal{L}_{\rm sp}$, to enlarge the stroke size. Thanks to the above key designs, we built an unconstrained resolution style transfer system for the first time.
	}
	\label{fig:overall_architecture} 
\end{figure*}

\section{Related Work}

\subsubsection{Neural Style Transfer.}
Inspired by the success of convolutional neural networks (CNNs), \cite{gatys2016image} first proposed a CNN-based style transfer algorithm, which opened up the new research field. 
To accelerate neural style transfer, \cite{johnson2016perceptual} and \cite{ulyanov2016texture} attempted to train a feed-forward network to learn a specific artistic style. 
In recent years, to improve the efficiency of transferring new styles, researchers have proposed many multiple style transfer~\cite{chen2017stylebank,dumoulin2017learned,li2017diversified,zhang2018multi} and arbitrary style transfer~\cite{gu2018arbitrary,huang2017arbitrary,deng2020arbitrary,lu2019closed, sheng2018avatar, yao2019attention} methods.
Nowadays, neural style transfer has achieved great progress, but due to massive memory cost and small stroke size, ultra-high resolution style transfer is still challenging.

\subsubsection{High-Resolution Neural Style Transfer.}
GPU memory is the main factor that restricts high-resolution style transfer. 
\cite{an2020real} proposed ArtNet, a lightweight network pruned from GoogLeNet~\cite{szegedy2015going} for neural style transfer.
\cite{jing2020dynamic} developed a MobileNet-based lightweight network, significantly reducing the computation complexities compared with the original VGG encoder. 
\cite{wang2020collaborative} proposed a distillation-based method, which used the pre-trained VGG19~\cite{simonyan2014very} as the teacher and a small encoder as the student, successfully rendering high-resolution images up to 6000$\times$6000 pixels on a single 12GB GPU. 
Although these methods reduce the memory consumption, they will still exhaust the GPU memory when processing ultra-high resolution images (\eg, 10000$\times$10000 pixels).

\subsubsection{Stroke Size Control in Neural Style Transfer.} 
Stroke size is an important perceptual factor highly related to the quality of style transfer results.
Typically, \emph{a stylized result with large brush strokes tends to have a better appearance than the one with small brush strokes.}
\cite{gatys2017controlling} first presented that the stroke size is related to the receptive field of the loss network,
and they proposed a coarse-to-fine method to generate stylized results with large brush strokes. 
\cite{wang2017multimodal} proposed a hierarchical network to enlarge the stroke size and trained it with multiple losses of increasing scales. 
\cite{jing2018stroke} presented a style-specific network with multiple stroke branches, supervised by multi-scale style images. 
\cite{zhang2018multi} developed a multi-style generative network (MSG-Net), which controls the stroke size by scaling the style image for inference. 
Nevertheless, these stroke size control methods are mainly designed for the style transfer under common image resolution~(\eg, 1000$\times$1000 pixels), which are difficult to apply in ultra-high resolution scenarios.

\begin{figure*}[ht]
	\centering
	\includegraphics[width=0.99\textwidth]{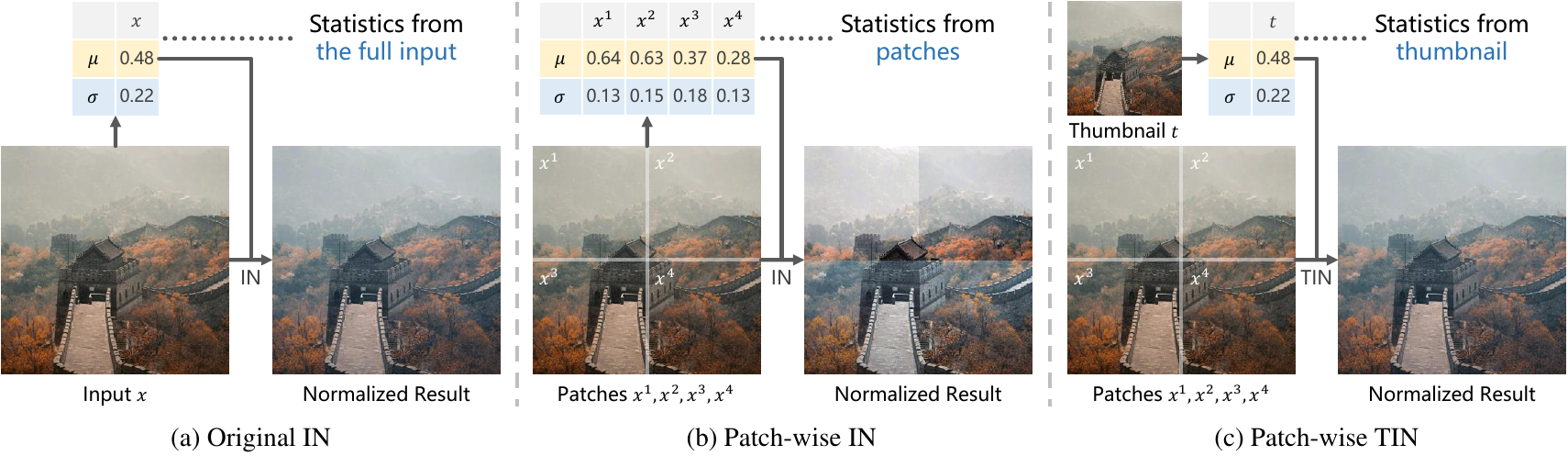}
	\caption{\textbf{A simple example of IN and the proposed TIN.} (a) We normalize the input as a whole. (b) We divide the input into four patches and normalize them individually. (c) We apply thumbnail's normalization statistics to these four patches, obtaining a similar output as (a). These results show that IN is not applicable to patch-wise style transfer.
	}
	\label{fig:in_tin_example} 
\end{figure*}

\section{Proposed Method}

\subsection{Overall Architecture}
\label{sec:overall_architecture}

The goal of the URST framework is to overcome the difficulties in GPU memory limit and small brush strokes when processing ultra-high resolution images. It consists of three key designs: 
(1) A flexible pipeline termed patch-wise style transfer that can convert a high-cost style transfer task to multiple low-cost patch stylization. 
(2) A novel thumbnail instance normalization (TIN) layer that can extract thumbnail features' normalization statistics and apply them to small patches, ensuring the style consistency among different patches.
(3) A carefully defined stroke perceptual loss that focuses on the perceptual differences in brush strokes, encouraging style transfer networks to keep large stroke size. 
Benefiting from these versatile designs, our URST can be easily plugged into most existing methods to perform ultra-high resolution style transfer.

An overview of the URST framework is depicted in Figure~\ref{fig:overall_architecture}.
Taking an ultra-high resolution content image $I_{\rm c}$ as input,
the pipeline of the URST can be divided into three stages: dividing, stylization, and assembling. 
(1) In the dividing stage, we first generate a thumbnail image $I_{\rm t}$ for each content image, and then divide the content image $I_{\rm c}$ into a sequence of small patches $\{I_{\rm p}^{i}\ |\ i=1,2,...,N \}$.
(2) In the stylization stage, the thumbnail image $I_{\rm t}$ is the first to be fed into the style transfer network, to collect the normalization statistics across the network.
Then, these normalization statistics are applying to stylize the small patches, obtaining the stylized patches $\{\hat{I}_{\rm p}^{i}\ |\ i=1,2,...,N \}$.
Here, our style transfer network is not specific. Most existing IN-based methods can be used as the style transfer network.
(3) In the assembling stage, all stylized patches are assembled into an ultra-high resolution stylized image $\hat{I}_{\rm c}$.

Since the style transfer network in our framework can be different methods (\eg, AdaIN~\cite{huang2017arbitrary} and LinearWCT~\cite{li2019learning}) whose loss functions are various, for convenience, we define the loss of the selected method as the original loss $\mathcal{L}_{\rm o}$.
During training, we first optimize the network with the original loss calculated on the stylized thumbnail.
In addition, URST introduces an auxiliary loss, termed stroke perceptual loss, to further improve the quality of ultra-high resolution style transfer. Its core idea is to penalize the perceptual differences in brush strokes between the stylized patch $\hat{I}_{\rm p}$ and the upsampled patch $\hat{I}_{\rm tp}$ that cropped from the stylized thumbnail $\hat{I}_{\rm t}$. It should be noticed that the upsampled patch $\hat{I}_{\rm tp}$ plays a role of the learning target. Therefore, the gradient flow of $\hat{I}_{\rm tp}$ is detached.

\subsection{Patch-wise Style Transfer}
To process ultra-high resolution images, we propose patch-wise style transfer. Given an ultra-high resolution content image $I_{\rm c}$, we use a $K\!\times\!K$ pixels sliding window with a stride of $S$ to divide the content image $I_{\rm c}$ into multiple overlapping patches $\{I_{\rm p}^{i}\ |\ i=1,2,...,N \}$.
Considering limited GPU memory resources, these patches will be fed to the network in batches. After the loop of patch stylization, we obtain a sequence of stylized patches $\{\hat{I}_{\rm p}^{i}\ |\ i=1,2,...,N \}$.
Finally, we discard the overlapping regions on these stylized patches and assemble them into a complete image $\hat{I}_{\rm c}$.

Compared with previous methods~\cite{an2020real, wang2020collaborative} that use a full image as input, this patch-wise manner can flexibly process arbitrary high-resolution images, and also be easily plugged into most existing style transfer methods, such as AdaIN~\cite{huang2017arbitrary}, WCT~\cite{li2017universal}, and LinearWCT~\cite{li2019learning}. However, it is evident that the style of stylized patches is inconsistent (see Figure~\ref{fig:comparison_linearwct}(c)), due to the individual normalization statistics calculated from each patch. Therefore, we propose the thumbnail instance normalization (TIN) to address this problem.

\begin{figure*}[ht]
	\centering
	\includegraphics[width=0.99\textwidth]{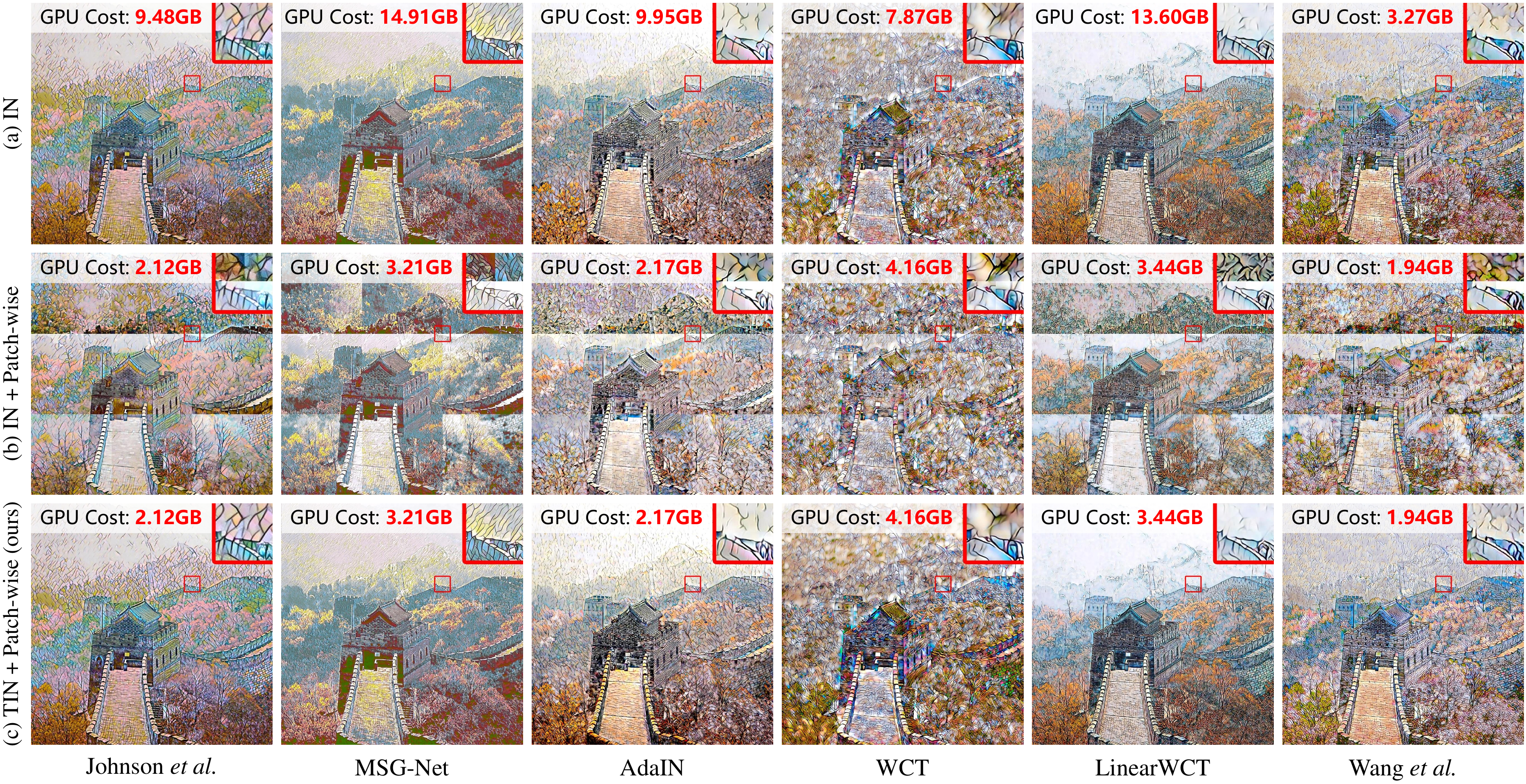}
	\caption{\textbf{Stylization comparison of different pipelines and normalization methods (3000$\times$3000 pixels).}
	To verify the versatility of our approach, we conduct experiments on 6 representative style transfer methods,
	incluing Johnson~\etal~\cite{johnson2016perceptual}, MSG-Net~\cite{zhang2018multi}, AdaIN~\cite{huang2017arbitrary}, WCT~\cite{li2017universal}, LinearWCT~\cite{li2019learning}, and Wang~\etal~\cite{wang2020collaborative}.
	The style image and content image are the same with Figure~\ref{fig:comparison_linearwct}.
	(a) shows the results directly generated by IN-based networks, which cost massive GPU memory. (b) and (c) show the results of patch-wise style transfer with IN and the proposed TIN, which cost much less memory than (a). Note that our results (c) are as high-quality as (a), which demonstrates the effectiveness of our TIN.
	}
	\label{fig:comparison_tin} 
\end{figure*}

\subsection{Thumbnail Instance Normalization}
\label{sec:tin}
IN is a widely-used normalization layer in neural style transfer. 
Given an input tensor $x\in\mathbb{R}^{N \times C \times H \times W}$, IN can be formulated as:
\begin{equation}
	\textrm{IN}(x)= \gamma\left(\frac{x-\mu(x)}{\sigma(x)}\right)+\beta,
\end{equation}
where $\mu(x), \sigma(x) \in \mathbb{R}^{N \times C}$ are channel-wise statistics; $\gamma, \beta \in \mathbb{R}^C$ are trainable affine parameters. However, we found that IN is not applicable to patch-wise style transfer, because stylized patches generated by the IN-based network are inconsistent in style. As demonstrated in Figure~\ref{fig:in_tin_example}, we take a simple example to illustrate this problem. In Figure~\ref{fig:in_tin_example}(a), we normalize the input as a whole. In Figure~\ref{fig:in_tin_example}(b), we divide the input into four patches and normalize them individually. Comparing these two results reveals that the underlying cause of style inconsistency is the individual normalization statistics calculated from each patch.

Based on the above analysis, we propose a simple variant to IN, termed thumbnail instance normalization (TIN). Our TIN receives a patch $x\in\mathbb{R}^{N \times C \times H \times W}$ and a thumbnail $t\in\mathbb{R}^{N \times C \times H_t \times W_t}$ as input, and it can be formulated as:
\begin{equation}
	\textrm{TIN}(x, t)= \gamma\left(\frac{x-\mu(t)}{\sigma(t)}\right)+\beta.
\end{equation}
Different from IN, here $\mu(t), \sigma(t) \in \mathbb{R}^{N \times C}$ are channel-wise mean and standard deviation of the thumbnail input $t$. In this way, our TIN is able to ensure the style consistency among different patches, as shown in Figure~\ref{fig:in_tin_example}(c).

Similarly, instance whitening (IW)~\cite{pan2019switchable} has the same problem, which is a standardization method based on second-order statistics (\ie, covariance matrix). It is also widely used in many neural style transfer methods~\cite{li2019learning, li2017universal, wang2020collaborative, yoo2019photorealistic}. Our TIN can be generalized to thumbnail instance whitening (TIW) with minor modifications. We will discuss this in the supplementary material.

\subsection{Stroke Perceptual Loss}

Based on the proposed TIN, we present an auxiliary loss for enlarging the stroke size, termed stroke perceptual loss:
\begin{equation}
	\mathcal{L}_{\rm sp}(\hat{I}_{\rm p},\hat{I}_{\rm tp})=\left \| \mathcal{F}_{l}(\hat{I}_{\rm p})-\mathcal{F}_l(\hat{I}_{\rm tp}) \right \|^2,
\end{equation}
where $\mathcal{F}_l$ is the output feature map of the layer $l$ in the VGG network. $\hat{I}_{\rm p}$ is a stylized patch with small brush strokes, and $\hat{I}_{\rm tp}$ is a stylized patch that cropped and upsampled from the stylized thumbnail $\hat{I}_{\rm t}$, which has large brush strokes.

Thanks to the proposed TIN, the input pair $(\hat{I}_{\rm p},\hat{I}_{\rm tp})$ has similar content and style, but the stroke size is different. 
Therefore, $\mathcal{L}_{\rm sp}$ can mainly measure the perceptual differences in brush strokes. 
In other words, optimizing this loss is to encourage the style transfer network to generate brush strokes as large as that of the target $\hat{I}_{\rm tp}$.

\subsection{Total Loss}
As mentioned above, we define the loss function used in the selected method as the original loss $\mathcal{L}_{\rm o}$. 
On this basis, we add the stroke perceptual loss $\mathcal{L}_{\rm sp}$ as an auxiliary loss. 
Therefore, the total loss is expressed as:
\begin{equation}
	\mathcal{L}=\mathcal{L}_{\rm o} + \lambda\mathcal{L}_{\rm sp},
	\label{total_loss}
\end{equation}
where $\lambda$ is a weight to balance $\mathcal{L}_{\rm o}$ and $\mathcal{L}_{\rm sp}$.
In our experiments, $\lambda$ is set to $1.0$ by default.

\section{Experiments}
\subsection{Implementation Details}

To verify the versatility of our URST, we apply it to 6 representative style transfer methods, including Johnson \etal~\cite{johnson2016perceptual}, MSG-Net \cite{zhang2018multi}, AdaIN \cite{huang2017arbitrary}, WCT \cite{li2017universal}, LinearWCT \cite{li2019learning}, and Wang \etal~\cite{wang2020collaborative}.

In the testing phase, we perform ultra-high resolution style transfer on photography works collected from \emph{pexels.com}. 
We use a 1064$\times$1064 pixels sliding window with a stride of 1000 to divide the input image, and the style image used in our framework is 1024$\times$1024 pixels. 
Besides, we scale the shorter side of the input image to 1024 pixels, as the thumbnail.

During training, our stroke perceptual loss is computed at the ${\tt relu4\_1}$ layer of the VGG network.
Following common practices~\cite{chen2016fast,deng2020arbitrary, li2019learning},
we use MS-COCO dataset~\cite{lin2014microsoft} as content images and WikiArt dataset~\cite{nichol2016wikiart} as style images, both of which contain roughly 80,000 training samples. 
Following previous methods,
we adopt a VGG19~\cite{simonyan2014very} pre-trained on ImageNet~\cite{deng2009imagenet} as the loss network.
All models are trained with a batch size of 8 on a Titan XP GPU, and other training settings are the same as the original settings in the selected style transfer methods~\cite{huang2017arbitrary, li2019learning}.

\subsection{Ablation Study}

\begin{figure*}[!ht]
	\centering
	\includegraphics[width=0.99\textwidth]{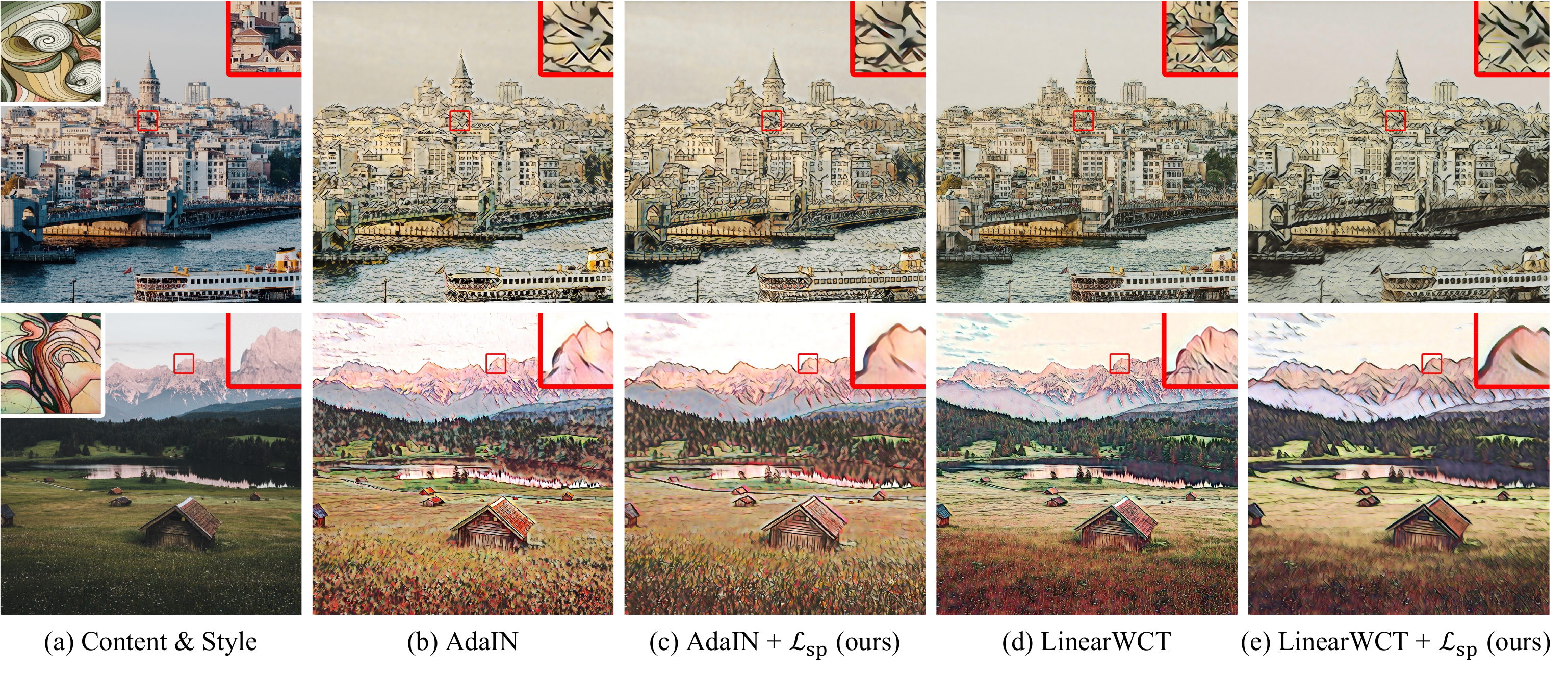}
	\caption{\textbf{Ablation study of the proposed stroke perceptual loss $\mathcal{L}_{\rm sp}$.} Comparison of these stylized images (3000$\times$3000 pixels) indicates that our $\mathcal{L}_{\rm sp}$ can significantly enlarge the stroke size of the existing style transfer methods.
	}
	\label{fig:comparison_stroke_size} 
\end{figure*}

\begin{figure}[!t]
	\centering
	\setlength{\fboxrule}{0pt}
	\fbox{\includegraphics[width=0.46\textwidth]{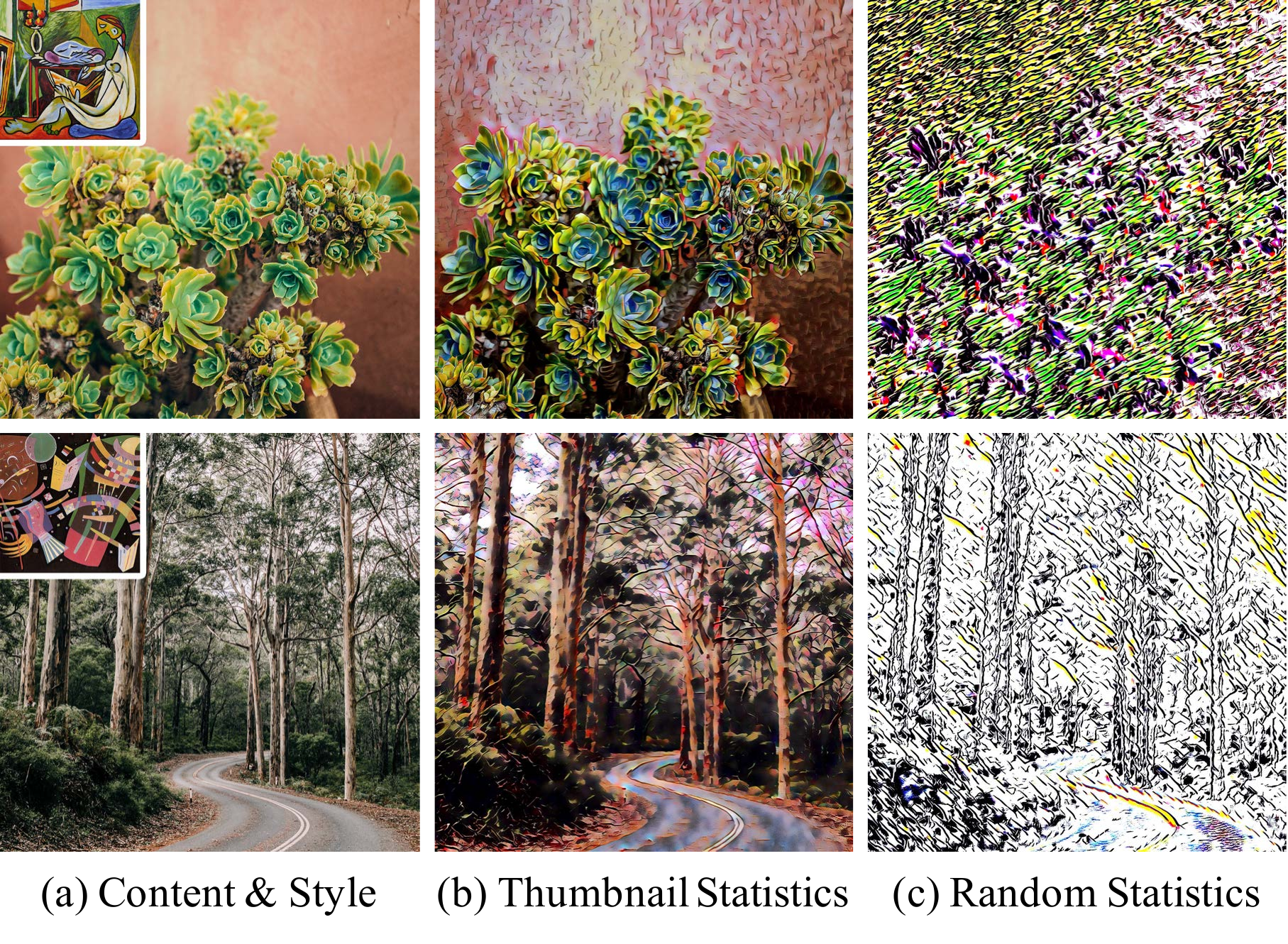}}
	\caption{
	\textbf{Thumbnail statistics \emph{vs.} random statistics.}
	This comparison demonstrates that using the normalization statistics of thumbnail features is the key to the success of patch-wise style transfer.}
	\label{fig:comparison_statistics} 
\end{figure}

\begin{figure}[!t]
	\centering
	\setlength{\fboxrule}{0pt}
	\fbox{\includegraphics[width=0.46\textwidth]{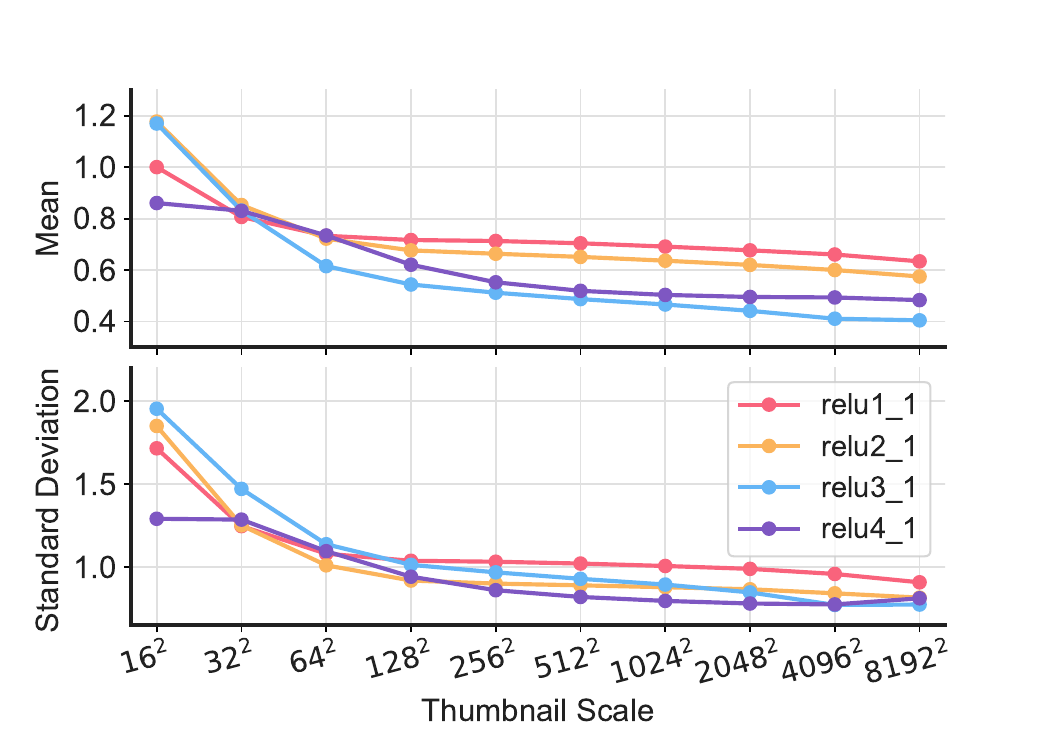}}
	\caption{\textbf{Mean and standard deviation of feature maps of the VGG19 network under different thumbnail scales.}
	It shows that with the growth of the thumbnail scale, the normalization statistics of feature maps tend to be stable.}
	\label{fig:vgg_statistics} 
\end{figure}

\subsubsection{Thumbnail Instance Normalization.}
As discussed, consistent normalization statistics are important for patch-wise style transfer. 
To verify this, we conduct experiments of patch-wise style transfer with IN and the proposed TIN, respectively. 
From Figure~\ref{fig:comparison_tin}(b), we can observe that IN leads to the style inconsistency among different patches. 
Differently, our method avoids this problem by adopting TIN (see Figure~\ref{fig:comparison_tin}(c)). 
In addition, we find that our results are as high-quality as the results demonstrated in Figure~\ref{fig:comparison_tin}(a), while our memory consumption is less than 5GB, showing that TIN can approximate the IN statistics of the original ultra-high resolution image, enabling the low memory cost ultra-high resolution style transfer.

Moreover, we compare the stylized results generated by the model with our TIN and the model with random normalization statistics in Figure~\ref{fig:comparison_statistics}.
Although using random normalization statistics can also keep the style consistency among different patches, it destroys the style information extracted from the style image, resulting in the unexpected styles as shown in Figure~\ref{fig:comparison_statistics}(c).
In contrast, using TIN not only ensures the style consistency among different patches, but also maintains the information of the target style.

\begin{figure*}[!t]
	\centering
	\includegraphics[width=0.99\textwidth]{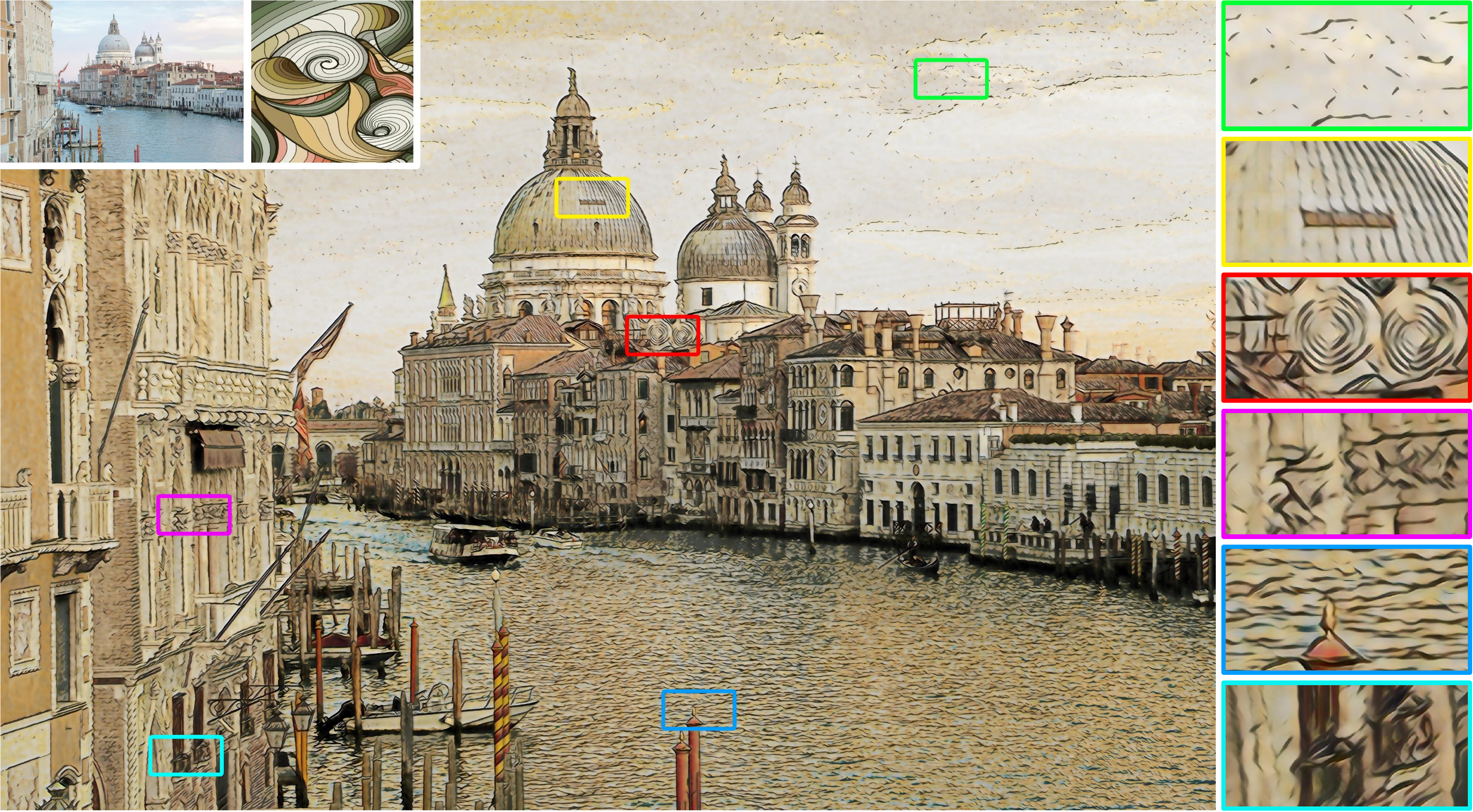}
	\caption{\textbf{An ultra-high resolution stylized result (12000$\times$8000 pixels)}, took about 2.5GB memory on a single 12GB GPU (Titan XP). On the upper left are the content image and style image. Six close-ups (660$\times$330 pixels) are shown on the right side of the stylized result.
	More ultra-high resolution stylized results are provided in the supplementary material.
	}
	\label{fig:ultra_high_result} 
\end{figure*}

\subsubsection{Thumbnail Size.}
To further study the relationship between normalization statistics and thumbnail size, 
we resize an ultra-high resolution image (8192$\times$8192 pixels) to the thumbnails of different scales, and calculate their normalization statistics in the style transfer network.
Specifically, we first feed these thumbnails to the encoder (\ie, VGG19) of the style transfer network and obtain the output feature maps of ${\tt relu1\_1}$, ${\tt relu2\_1}$, ${\tt relu3\_1}$, and ${\tt relu4\_1}$.
Then, we calculate the mean and standard deviation of these feature maps, and 
plot them
in Figure~\ref{fig:vgg_statistics}. 
Note that when the thumbnail scale is equal to 8192$\times$8192, the normalization statistics is the IN statistics.
We see that with the growth of the thumbnail scale, the normalization statistics of feature maps tend to be stable. 
When the thumbnail scale is larger than 1024$\times$1024 pixels, the TIN statistics are very close to IN statistics.
This indicates that TIN can well approximate IN when the thumbnail scale is larger than 1024$\times$1024.
In addition, we also conduct a qualitative ablation study for the thumbnail size in the supplementary material, from which the same conclusion can be drawn.
To balance speed and style transfer quality, we set the shorter side of the thumbnail to 1024 pixels by default.

\subsubsection{Stroke Perceptual Loss.} 
As shown in Figure~\ref{fig:comparison_stroke_size}, using the proposed stroke perceptual loss $\mathcal{L}_{\rm sp}$ as an auxiliary loss for neural style transfer can significantly enlarge the stroke size of these existing methods. 
Compared with the baseline results (see Figure~\ref{fig:comparison_stroke_size}(b)(d)), with the guidance of $\mathcal{L}_{\rm sp}$, these models learned to use thicker lines and sparser textures to depict the scenery, which helps to improve the quality of ultra-high resolution style transfer (see Figure~\ref{fig:comparison_stroke_size}(c)(e)).

\subsection{Discussion}
Different from previous methods~\cite{an2020real, wang2020collaborative}, URST is a versatile framework that can be easily plugged into most existing IN/IW-based methods.
Moreover, with the growth of the input resolution, its GPU memory overhead hardly increases. 
Theoretically, URST supports style transfer of arbitrary resolution images.

To further demonstrate the effectiveness of URST, we evaluate it on an ultra-high resolution image of 12000$\times$8000 pixels (\ie, 96 megapixels), as shown in Figure~\ref{fig:ultra_high_result}.
This result is produced based on AdaIN \cite{huang2017arbitrary} and only costs 2.5GB of GPU memory.
It also shows that our URST has achieved superior performance in producing large brush strokes due to the effectiveness of the stroke perceptual loss.
In conclusion, to our knowledge, this is the first time to build an unconstrained resolution style transfer system on a single 12GB GPU (Titan XP). 

\subsection{Limitation}
One limitation of URST is that it can not be applied to the optimization-based methods, such as Gatys \etal~\cite{gatys2016image} and STROTSS~\cite{kolkin2019style},
because these methods do not adopt IN \cite{ulyanov2016instance} or IW \cite{pan2019switchable}.
But we think that ``containing IN or IW" is a loose premise, since most existing style transfer methods can meet this prerequisite. 
In addition, we'd like to point out that these optimization-based methods are relatively slow, which always take more than 100 seconds to stylize an image of 1000$\times$1000 pixels, are not the best choice for ultra-high resolution style transfer.

\section{Conclusion}
In this work, we propose URST, a simple yet effective framework for arbitrary high-resolution style transfer. 
We perform patch-wise style transfer to process ultra-high resolution input under limited GPU memory resources, and develop a thumbnail instance normalization (TIN) layer to ensure the style consistency among different patches. 
Moreover, to enlarge the brush strokes in ultra-high resolution stylized results, the stroke perceptual loss $\mathcal{L}_{\rm sp}$ is introduced as an auxiliary loss for neural style transfer. 
Extensive experiments show that our URST surpasses existing SOTA methods on ultra-high resolution images and can be easily plugged into most existing IN/IW-based methods.
Although we mainly study neural style transfer in this work, instance normalization is also widely used in other low-level vision tasks. Therefore, the application of our TIN on other tasks is worth exploring in the future.


\newpage
{
\bibliographystyle{aaai22}
\bibliography{arxiv}
}

\newpage

\section{Appendix}

\subsection{Thumbnail Instance Whitening}
In recent years, instance whitening (IW)~\cite{pan2019switchable} has been a widely-used module in neural style transfer~\cite{li2019learning, li2017universal, wang2020collaborative, yoo2019photorealistic}. Instead of normalizing data using channel-wise mean and
standard deviation like instance normalization (IN)~\cite{ulyanov2016instance}, IW whitens the data using its covariance matrix, which gives rise to better stylization quality than IN. Given an input tensor $x\in\mathbb{R}^{N \times C \times H \times W}$, IW can be formulated as:
\begin{equation}\label{equ:iw}
	\textrm{IW}(x)= [\Sigma(x)]^{-\frac{1}{2}}(x-\mu(x)),
\end{equation}
where $\Sigma(x)\in \mathbb{R}^{N \times C\times C}$ and $\mu(x)\in \mathbb{R}^{N \times C}$ are covariance matrices and mean vectors.

However, IW has the same problem as IN when performing patch-wise style transfer. As discussed, the style of stylized patches is inconsistent due to the individual statistics calculated from each patch. To address this problem, we generalize our TIN to its IW version, namely thumbnail instance whitening (TIW). As same as TIN, our TIW receives a patch $x\in\mathbb{R}^{N \times C \times H \times W}$ and a thumbnail $t\in\mathbb{R}^{N \times C \times H_t \times W_t}$ as input, and it can be formulated as:
\begin{equation}\label{equ:tiw}
	\textrm{TIW}(x,t)= [\Sigma(t)]^{-\frac{1}{2}}(x-\mu(t)),
\end{equation}
where $\Sigma(t)\in \mathbb{R}^{N \times C\times C}$ and $\mu(t)\in \mathbb{R}^{N \times C}$ are covariance matrices and mean vectors of the thumbnail input $t$. 
In this way, we avoid the style inconsistency among different patches for most existing style transfer methods.

\begin{table}[!b]
	\centering
	\scriptsize
	\caption{\textbf{GPU memory comparison (in GB).}
	All results are tested by PyTorch~\cite{paszke2019pytorch} on a Titan XP GPU (12GB). ``$-$" denotes out-of-memory.}
	\setlength{\tabcolsep}{0.6mm}{
	
		\begin{tabular}{@{\extracolsep{\fill}}ccccccccccccc} \toprule
			\multirow{2}{*}{\textbf{Res.}} &\multicolumn{2}{c}{\textbf{Johnson}}& \multicolumn{2}{c}{\textbf{MSG-Net}} & \multicolumn{2}{c}{\textbf{AdaIN}}& \multicolumn{2}{c}{\textbf{WCT}}& \multicolumn{2}{c}{\textbf{LinearWCT}} & \multicolumn{2}{c}{\textbf{Wang~\etal}}\\ 
			& Orig. & Ours &  Orig. & Ours & Orig. & Ours & Orig. & Ours & Orig. & Ours & Orig. & Ours\\
			\midrule
			1000$^2$  & 1.21  & 2.12  & 2.55 & 3.21  & 1.40 & 2.17  & 1.84  & 4.61  & 2.79 & 3.44  & 0.90  & 1.94  \\ 
			2000$^2$  & 4.31  & 2.12  & 7.15 & 3.21  & 4.62 & 2.17  & 4.10  & 4.61  & 6.85 & 3.44  & 1.80  & 1.94  \\
			3000$^2$  & 9.48  & 2.12  & $-$  & 3.21  & 9.95 & 2.17  & 7.87  & 4.61  & $-$  & 3.44  & 3.27  & 1.94  \\
			4000$^2$  & $-$   & 2.12  & $-$  & 3.21  & $-$  & 2.17  & $-$   & 4.61  & $-$  & 3.44  & 5.36  & 1.94  \\
			5000$^2$  & $-$   & 2.12  & $-$  & 3.21  & $-$  & 2.17  & $-$   & 4.61  & $-$  & 3.44  & 8.00  & 1.94  \\
			6000$^2$  & $-$   & 2.12  & $-$  & 3.21  & $-$  & 2.17  & $-$   & 4.61  & $-$  & 3.44  & 11.28 & 1.94  \\
			7000$^2$  & $-$   & 2.12  & $-$  & 3.21  & $-$  & 2.17  & $-$   & 4.61  & $-$  & 3.44  & $-$   & 1.94 \\
			8000$^2$  & $-$   & 2.12  & $-$  & 3.21  & $-$  & 2.17  & $-$   & 4.61  & $-$  & 3.44  & $-$   & 1.94 \\
			9000$^2$  & $-$   & 2.12  & $-$  & 3.21  & $-$  & 2.17  & $-$   & 4.61  & $-$  & 3.44  & $-$   & 1.94 \\
			10000$^2$ & $-$   & 2.12  & $-$  & 3.21  & $-$  & 2.17  & $-$   & 4.61  & $-$  & 3.44  & $-$   & 1.94 \\
			\bottomrule
		\end{tabular}
	}
	\label{tab1}
\end{table}

\subsection{Memory and Speed Analysis}
\subsubsection{GPU Memory Usage.}
In this section, we report the memory cost for a single content image of different resolutions. 
As shown in Table~\ref{tab1}, most existing methods cannot process high-resolution images (\eg, 4000$\times$4000 pixels) with limited memory.
\cite{wang2020collaborative} is a recent distillation-based method designed for high-resolution style transfer, but it can only process up to 6000$\times$6000 pixels.
Unlike them, our URST can keep the memory cost below 5GB, and with the growth of the input resolution, our GPU memory overhead hardly increases. 
Theoretically, URST supports style transfer of arbitrary resolution images.

\subsubsection{Time Cost.}
The time cost of URST is linear to the number of patches.
For example, on a single Titan XP GPU, AdaIN \cite{huang2017arbitrary} takes about 0.15 seconds to stylize a 1000$\times$1000 image, thus ``URST + AdaIN" takes about 15 seconds to process a 10000$\times$10000 image.

\subsection{More Applications}
\subsubsection{Image-to-Image Translation.}
\begin{figure}[h]
    \vspace{-0.1cm}
    \setlength{\abovecaptionskip}{0.1cm}
	\centering
	\includegraphics[width=0.47\textwidth]{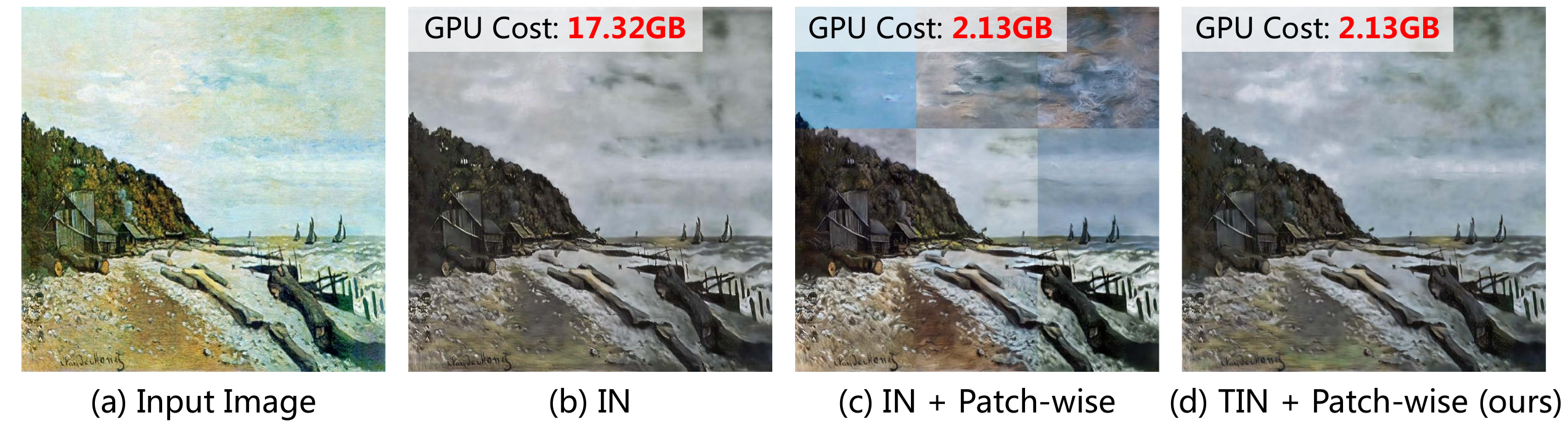}
	\caption{\textbf{Experiments on CycleGAN {(monet-to-photo task)}.}
	The experimental settings are the same as Figure \ref{fig:comparison_tin}.
	}
	\label{fig_a} 
	\vspace{-0.1cm}
\end{figure}

To further verify the effectiveness of our method, we conduct an extra experiment on CycleGAN \cite{zhu2017unpaired}, a representative image-to-image translation method. 
Specifically, we adopt its official code\footnote{\url{https://github.com/junyanz/pytorch-CycleGAN-and-pix2pix}} and pre-trained models, and replace all IN layers with our TIN layers.
As shown in Figure \ref{fig_a}, our TIN can also be successfully applied on CycleGAN.

\subsubsection{Flexible User Control.}
\begin{figure}[h]
    \vspace{-0.1cm}
    \setlength{\abovecaptionskip}{0.1cm}
	\centering
	\includegraphics[width=0.47\textwidth]{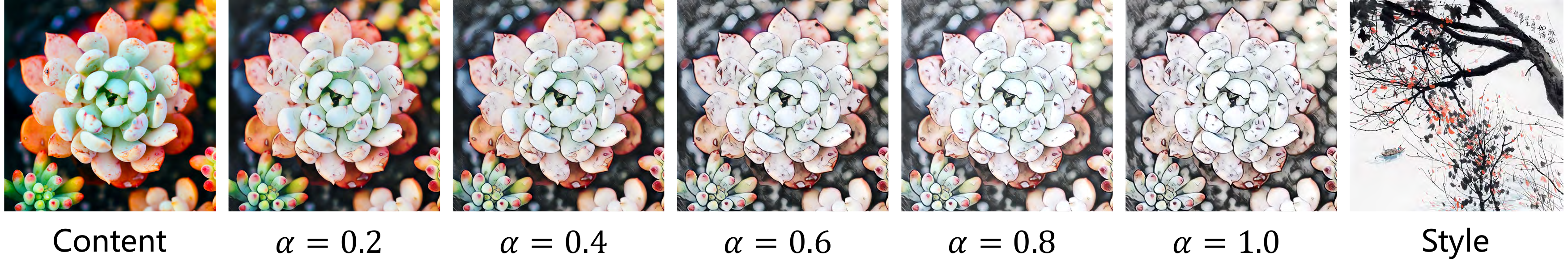}
	\caption{\textbf{Trade-off between the content and style image.}
	These results are based on LinearWCT~\cite{li2019learning}.
	}
	\label{fig_c} 
	\vspace{-0.1cm}
\end{figure}

Our URST is compatible with various user controls. Here, we take \emph{style weight control} as an example, as shown in Figure \ref{fig_c}.

\subsection{More Experimental Results}
\subsubsection{Qualitative Ablation for Thumbnail Size.} 
\begin{figure}[h]
    \vspace{-0.1cm}
    \setlength{\abovecaptionskip}{0.1cm}
	\centering
	\includegraphics[width=0.47\textwidth]{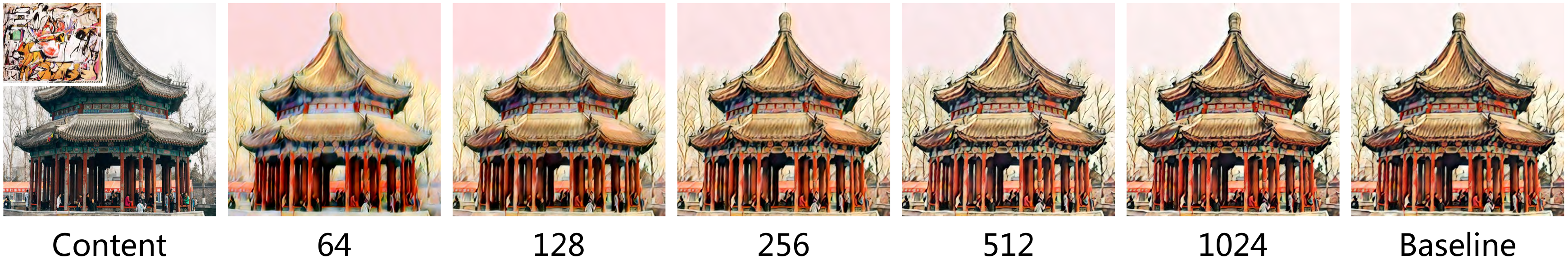}
	\caption{\textbf{Qualitative results with different thumbnail sizes.}
	These results are based on LinearWCT~\cite{li2019learning}.
	}
	\vspace{-0.1cm}
	\label{fig_b} 
\end{figure}

As presented in Figure \ref{fig_b}, as the thumbnail size increases, the result is gradually close to baseline's style transfer performance, showing that TIN can approximate IN statistics of ultra high-resolution images when the thumbnail size reaches 1024.

\subsubsection{More Ultra-high Resolution Results.}

\begin{figure*}[t]
	\centering
	\subfigure[URST + AdaIN~\cite{huang2017arbitrary}]{
		\includegraphics[width=\textwidth]{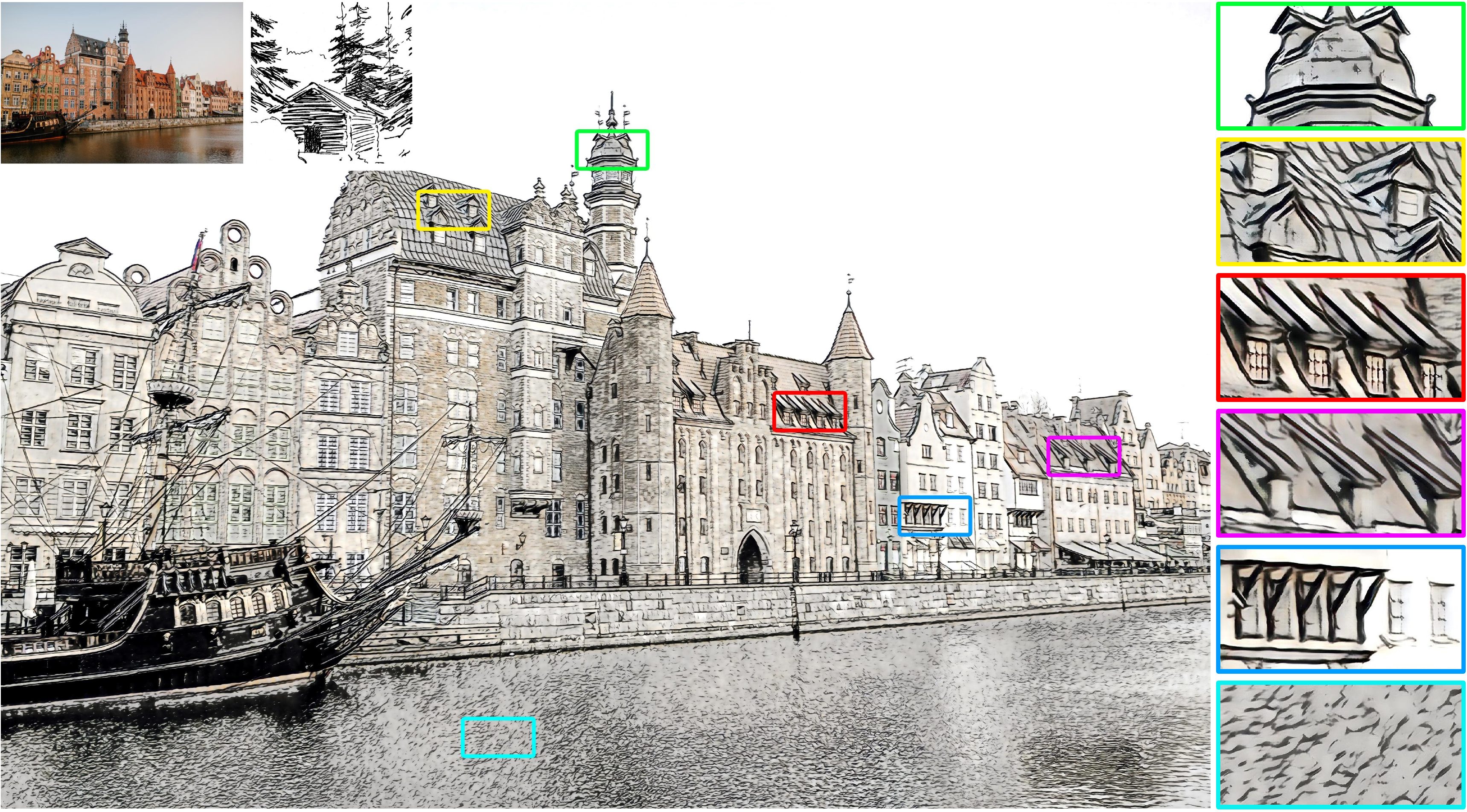}
	}
	\quad
	\subfigure[URST + LinearWCT~\cite{li2019learning}]{
		\includegraphics[width=\textwidth]{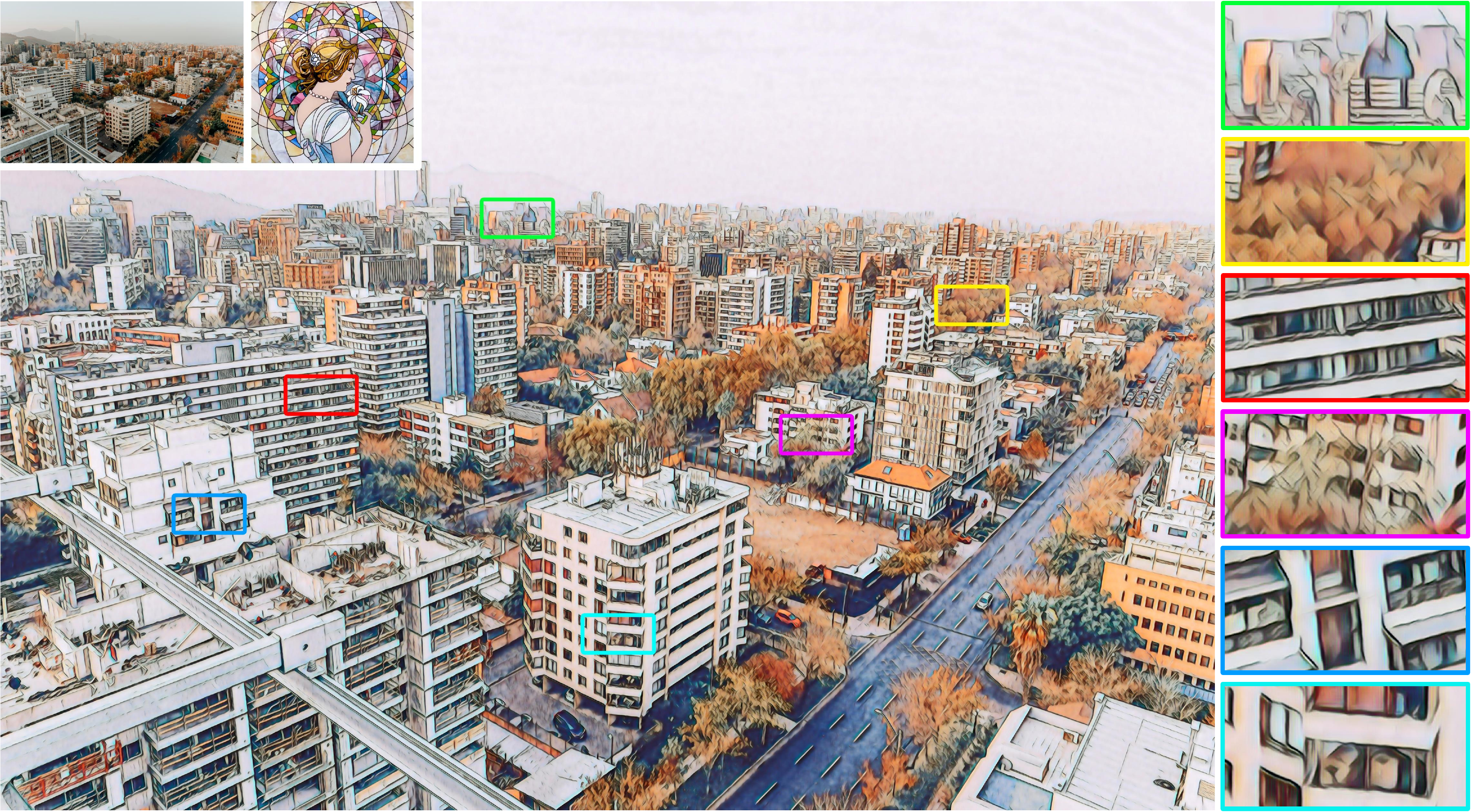}
	}
	\caption{\textbf{More ultra-high resolution stylized results (12000$\times$8000 pixels) produced on a single 12GB GPU (Titan XP).} On the upper left are the content image and style image. Six close-ups (660$\times$330 pixels) are shown on the right side of the stylized result.}
	\label{fig:ultra_high_result2}
\end{figure*}

More ultra-high resolution results of 12000$\times$8000 pixels (\ie, 96 megapixels) produced by our URST framework are presented in Figure~\ref{fig:ultra_high_result2},
which are based on AdaIN~\cite{huang2017arbitrary} and LinearWCT~\cite{li2019learning}, respectively.

\end{document}